%% file: arxiv.tex
\documentclass[10pt,twocolumn,letterpaper]{article}

\usepackage{iccv}
\usepackage{times}
\usepackage{epsfig}
\usepackage{graphicx}
\usepackage{amsmath}
\usepackage{amssymb}

\usepackage{booktabs}
\usepackage{algpseudocode}
\usepackage{mathtools}
\usepackage{listings}
\usepackage{xcolor}
\usepackage{siunitx}
\usepackage{multirow}
\usepackage{xspace}
\usepackage{enumitem}
\usepackage{ulem}
\usepackage{subcaption}
\graphicspath{ {./images/} }

\definecolor{codegreen}{rgb}{0,0.6,0}
\definecolor{codegray}{rgb}{0.5,0.5,0.5}
\definecolor{codepurple}{rgb}{0.58,0,0.82}
\definecolor{backcolour}{rgb}{0.95,0.95,0.92}

\lstdefinestyle{mystyle}{
    backgroundcolor=\color{backcolour},   
    commentstyle=\color{codegreen},
    keywordstyle=\color{magenta},
    numberstyle=\tiny\color{codegray},
    stringstyle=\color{codepurple},
    basicstyle=\ttfamily\footnotesize,
    breakatwhitespace=false,         
    breaklines=true,                 
    captionpos=b,                    
    keepspaces=true,                 
    numbers=left,                    
    numbersep=5pt,                  
    showspaces=false,                
    showstringspaces=false,
    inputencoding=utf8,
    extendedchars=false,
    showtabs=false,                  
    tabsize=2
}

\usepackage[pagebackref=true,breaklinks=true,letterpaper=true,colorlinks,bookmarks=false]{hyperref}

\usepackage[capitalize]{cleveref}
\crefname{section}{Sec.}{Secs.}
\Crefname{section}{Section}{Sections}
\Crefname{table}{Table}{Tables}
\crefname{table}{Tab.}{Tabs.}
\crefname{figure}{Fig.}{Fig.}

\iccvfinalcopy 

\def\iccvPaperID{9089} 

\ificcvfinal\fi

\newcommand{\model}{SiLK\xspace}

\newcommand{\VGG}{VGGnp}
\begin{document}

\title{\model : Simple Learned Keypoints}
\input{authors}
\maketitle

\ificcvfinal\thispagestyle{empty}\fi

\input{egpaper_core}

\clearpage

\pagebreak

\appendix

\noindent {\Huge \textbf{Appendix}}
\newline


\input{suppl/math}
\input{suppl/experiments}

\input{suppl/implementation}



\end{document}

%% file: authors.tex
\author{Pierre Gleize\\
{\tt\small gleize@meta.com}
\and
Weiyao Wang\\
{\tt\small weiyaowang@meta.com}\\
\\Meta AI
\\\url{https://github.com/facebookresearch/silk}
\and
Matt Feiszli\\
{\tt\small mdf@meta.com}
\\
}

%% file: egpaper_core.tex
\input{sections/abstract}

\input{sections/intro}

\input{sections/related_work}

\input{sections/methodology}

\input{sections/experiments}
\input{sections/analysis}

\input{sections/conclusion}

{\small
\bibliographystyle{ieee_fullname}
\bibliography{egbib}
}

%% file: sections/abstract.tex
\begin{abstract}

    Keypoint detection \& descriptors are foundational technologies for computer vision tasks like image matching, 3D reconstruction and visual odometry. Hand-engineered methods like Harris corners, SIFT, and HOG descriptors have been used for decades; more recently, there has been a trend to introduce learning in an attempt to improve keypoint detectors. 
    On inspection however, the results are difficult to interpret; recent learning-based methods employ a vast diversity of experimental setups and design choices: empirical results are often reported using different backbones, protocols, datasets, types of supervisions or tasks. Since these differences are often coupled together, it raises a natural question on what makes a good learned keypoint detector.
    In this work, we revisit the design of existing keypoint detectors by deconstructing their methodologies and identifying the key components. We re-design each component from first-principle and propose \textbf{Si}mple \textbf{L}earned \textbf{K}eypoints (SiLK) that is fully-differentiable, lightweight, and flexible. Despite its simplicity, SiLK advances new state-of-the-art on Detection Repeatability and Homography Estimation tasks on HPatches and 3D Point-Cloud Registration task on ScanNet, and achieves competitive performance to state-of-the-art on camera pose estimation in 2022 Image Matching Challenge and ScanNet.
    
    
\end{abstract}

%% file: sections/intro.tex
\section{Introduction}
\label{sec:intro}

\begin{figure}[h]
\centering
\includegraphics[width=\linewidth]{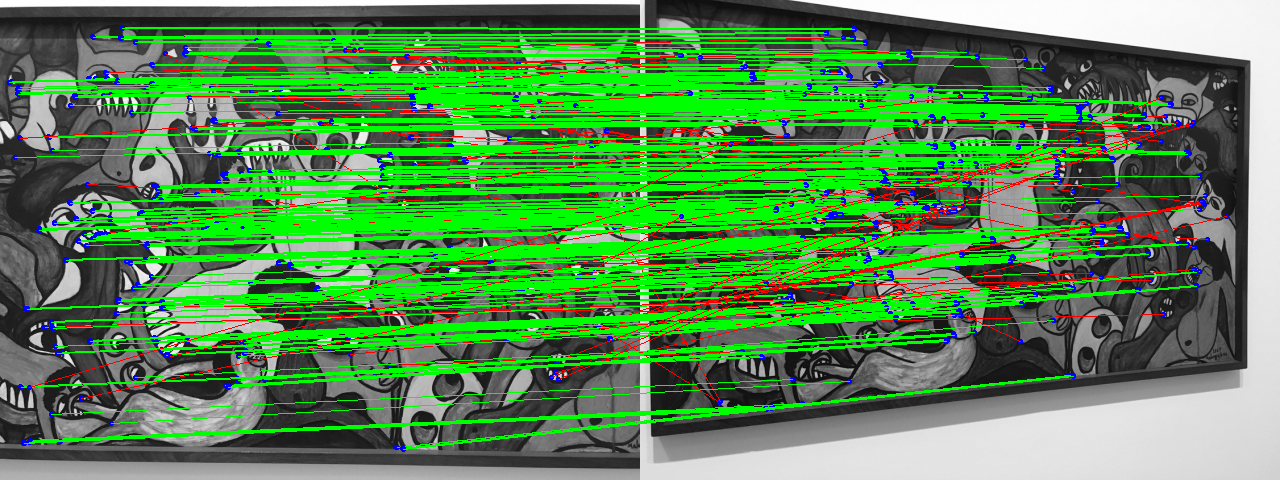}
\includegraphics[width=\linewidth]{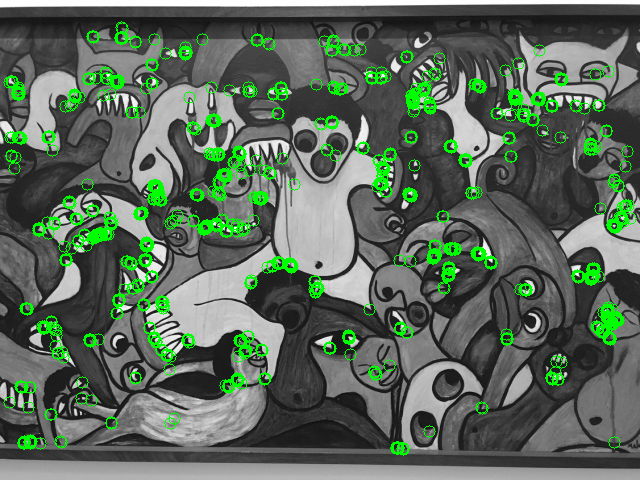}
\label{fig:arch}
\caption{The top image is an example of keypoint matching under viewpoint change; correct matches are green, while incorrect ones are red. The bottom image shows keypoints which are cycle-consistent by \model. As can be observed, \model has learned to find distinctive geometric features (corners, curves, intersections,..); from single non-annotated images.}
\end{figure}

Keypoint detection and matching is a foundational computer vision technique to obtain a sparse yet informative representation of an image or video sequence.  
Image stitching~\cite{Brown2006AutomaticPI,adel2014image}, SLAM~\cite{Davison2007MonoSLAMRS,MurArtal2015ORBSLAMAV}, SfM~\cite{schonberger2016structure}, camera calibrations, tracking~\cite{10.1007/978-3-030-58565-5_3}, and object detection~\cite{10.1007/978-3-319-46448-0_2} are important tasks which have been built on keypoint correspondences~\cite{ma2021image}. 
For a given task, a good keypoint representation should be able to identify a small set of points which are useful and informative for the task.  One typically also wants robustness of the descriptor to some set of transformations, like scale, point of view, or lighting variation.  In this writeup we are primarily interested in keypoints for 3D geometric problems like SLAM and SfM, which rely on finding correspondences between pairs of images.  For such problems, good keypoints should contain enough information to maximize the probability of correct matching from different points of view in varying conditions.

Our contributions are twofold.  
\begin{enumerate}[noitemsep]
    \item After reviewing a number of alternative approaches we propose Simple Learned Keypoints (\model), designed to be the simplest possible self-supervised approach to learn distinctive and robust keypoints from arbitrary image data in a traditional ``detect-and-describe" framework.  Despite its simplicity, \model is competitive or surpasses SOTA in most settings. 
    \item Leveraging \model's simple one-stage training protocol and modular architecture, we are able to ablate various dimensions of detector performance for different tasks.  In particular, with an eye toward real-time performance, we identify tasks where extremely lightweight backbone architectures are sufficient.
\end{enumerate}

\begin{table*}
  \setlength{\tabcolsep}{1.5pt}
  \centering
  \footnotesize
  \begin{tabular}{ l|ccccc|ccc|cccc}
 \hline
 & \multicolumn{5}{c|}{Keypoint Detection} & \multicolumn{3}{c|}{Descriptors \& Matching} & \multicolumn{3}{c}{Model \& Training} \\
 \hline
 & Learned & Sparse & Cell-based & NMS & Supervision & CA & Matcher & Supervision & Input & Backbone & Data & E2E \\
 \hline
 SIFT & No & {\checkmark} & \textbf{No} & \checkmark & - & \textbf{No} & \textbf{MNN} & - & - & - & - & - \\
 \hline \hline
 SuperPoint & \pmb{\checkmark} & {\checkmark} & \checkmark & \checkmark & Homo. Adapt. & \textbf{No} & \textbf{MNN} & \textbf{Rand. Homo.} & \textbf{Grey} & VGG & COCO & No \\
 SuperGlue & - & {\checkmark} & - & - & - & GraphNN & OT & SfM &  \textbf{Grey} & VGG & Oxford \& Paris & No\\
 GLAMPoints & \pmb{\checkmark} & {\checkmark} & \textbf{No} & \checkmark & \textbf{Matching Succ.} & \textbf{No} & \textbf{MNN} & - & \textbf{Grey} & UNet & SlitLamp & \pmb{\checkmark}  \\
 D2-Net & \pmb{\checkmark} & {\checkmark} & \checkmark & \checkmark & Triplet Ranking & \textbf{No} & \textbf{MNN} & SfM & RGB & VGG & MegaDepth & \pmb{\checkmark}\\
 R2D2 & \pmb{\checkmark} & {\checkmark} & \checkmark & \checkmark & Matching AP & \textbf{No} & \textbf{MNN} & SfM & RGB & L2-Net & Aachen & \pmb{\checkmark}\\
 DISK & \pmb{\checkmark} & {\checkmark} & \checkmark & \checkmark & \textbf{Matching Succ.} & \textbf{No} & \textbf{MNN} & SfM & RGB & UNet & MegaDepth & \pmb{\checkmark} \\
 URR & - & No & - & - & - & - & \textbf{MNN} & 3D Rendering & RGB & ResNet & ScanNet & \pmb{\checkmark}\\
 LoFTR & - & No & - & - & - & Transformer &  \textbf{MNN},OT & SfM & \textbf{Grey} & FPN & ScanNet,MegaDepth & \pmb{\checkmark} \\
 \hline
 \hline
 \model & \pmb{\checkmark} & \textbf{Optional} & \textbf{No} & \textbf{No} & \textbf{Matching Succ.} & \textbf{No} & \textbf{MNN} & \textbf{Rand. Homo.} & \textbf{Grey} & \textbf{Generic} & \textbf{Any Image Set} & \pmb{\checkmark} \\
 \hline
 \end{tabular}
 \caption{\textbf{Non-exhaustive deconstruction of keypoint detectors along different dimensions.} On each, \model adopts the simplest choice or is flexible to different choices. In particular, \model does not depend strongly on backbone type or training data (\cref{tab:ana-backbone-results} \& \cref{tab:ana-dataset}).}
 \label{tab:key-det} \label{tab:model-learning}
\end{table*}

%% file: sections/related_work.tex
\section{Related Work}
\label{sec:related_work}


Early work focused on carefully engineered methods to identify distinctive keypoints with descriptors which are robust to changes such as viewpoint and illumination. Hand-crafted techniques like Harris-corners\cite{harris1988combined}, SIFT\cite{lowe2004distinctive}, ORB\cite{rublee2011orb} and others \cite{rosten2006machine,calonder2010brief,leutenegger2011brisk,alahi2012freak,bay2008speeded,dalal2005histograms} have used explicit geometric notions like corners, gradients, and scale-space extrema to achieve results which remain both efficient and competitive to date \cite{schonberger2016structure,MurArtal2015ORBSLAMAV}.


More recent work like SuperPoint\cite{detone2018superpoint} chose to learn to find corners; they generate a large set of synthetic shapes with annotated corners and train a model with this ground truth.  While providing compelling evidence for learned methods, their training procedure is quite complex: it contains multiple training phases, a synthetic dataset, and employs a homography adaptation trick that can be difficult to tune (see our reproduced results in \cref{tab:exp-hpatches-sparse}).


In the same spirit as more recent work \cite{dusmanu2019d2,revaud2019r2d2,tyszkiewicz2020disk,truong2019glampoints,hartmann2014predicting,christiansen2019unsuperpoint,tang2019neural}, \model aims to learn keypoints in simple end-to-end fashion, without explicitly defining them as corners.

Several attempts have been made to learn keypoints implicitly; either by the careful design of the loss \cite{dusmanu2019d2,revaud2019r2d2,christiansen2019unsuperpoint,tang2019neural}; or by directly predicting the matching success of descriptors \cite{tyszkiewicz2020disk,hartmann2014predicting,truong2019glampoints}. \model falls in the second category, but with slight twist (cf. \cref{sec:meth-keypoint-learning}).


To learn descriptors, contrastive losses are commonly used. Similar to \cite{tyszkiewicz2020disk,sun2021loftr}, \model adopts a probabilistic approach by modeling the matching probabilities in a double-softmax, cycle-consistency setting and optimizes the log likelihood. The probabilistic formulation, similar to InfoNCE \cite{oord2018representation}, gives us a clean way to reason about matching, and the abundant supply of hard negative examples (from pixels in the same image) makes it an attractive choice.

Context aggregation (CA) is a recent addition to the toolkit. Initiated by SuperGlue\cite{sarlin2020superglue}, CA aims to refine or transform descriptors from a \textit{pair} of images before matching them. Implemented as a GNN\cite{scarselli2008graph} in \cite{sarlin2020superglue}, or as a Transformer\cite{vaswani2017attention} in \cite{sun2021loftr}, CA's predictions are conditional on all descriptors from the pair of images. In other words, for an image, the descriptors will be different when matching against different images. As a result, CA needs to run on every pair of images prior to matching, as opposed to running on single images. The run-time implications render CA prohibitively expensive in some applications (quadratic versus linear complexity). \model does not use CA, but outperforms \cite{sarlin2020superglue} and performs competitively with \cite{sun2021loftr}. Incorporating CA is optional for future works if performance is paramount and computation cost is less of a concern.


As a postprocessing after CA, SuperGlue\cite{sarlin2020superglue} introduced the concept of differentiable optimal transport (OT) to improve matching, using the Sinkhorn algorithm\cite{sinkhorn1967concerning}. LoFTR \cite{sun2021loftr} leveraged OT as well, but found little difference between OT and the simpler approach of mutual nearest neighbor (MNN) in some benchmarks.

%% file: sections/methodology.tex
\section{Methodology}
\label{sec:methodology}



\model's contribution is simplicity and flexibility. Our solution is built on the traditional approach of identifying distinctive pixels via robust local descriptors. We use modern but established techniques to learn to localize and describe keypoints given an arbitrary source of unlabeled images. Unlike classical methods, our descriptors and invariances are learned, and unlike some modern methods, there is no particular complexity in the matching process (\model employs only cosine distances and mutual nearest neighbor); this leaves few structural hyperparameters to tune. The simple backbone+heads design is backbone-agnostic, allowing experimentation. The annotation-free SSL approach means \model can be trained on any image or video dataset. Finally, a simple, one-stage training pipeline allows us to easily train and ablate different architectures, datasets, and hyperparameters for different tasks.

\model is trained to identify keypoints from single grayscale images. It provides both keypoint detections (location) and keypoint descriptors (for matching). Cycle consistency is employed for descriptor learning and a binary classifier identifies distinctive keypoints at pixel-level.

To learn descriptors, we take a source image and a transformed copy, extract descriptors for each point, and use descriptor similarity to define transition probabilities from a source location to each transformed location (and vice-versa). We optimize the descriptors to maximize cycle-consistency; i.e. we maximize the probability of a round-trip from the source to its transformed location and back.

To locate good keypoints, we train a binary classifier to identify points which will satisfy a matching criterion. A point and its transformed copy are positives when they are mutual nearest neighbors in the sense of transition probabilities, and they are negatives otherwise. We train both the cycle-consistency and classification losses jointly. 

We provide simple pseudo-code in \cref{fig:algorithm}.

\begin{figure}[h]
\centering
\begin{lstlisting}[language=Python]
def loss(image_0, model):
    # get warped image and pixel correspondences
    image_1, corr_0, corr_1 = rand_homo(image_0)
    
    # apply image augmentations
    image_0 = augment(image_0)
    image_1 = augment(image_1)
    
    # extract dense descriptors and keypoints
    desc_0, kpts_0 = model(image_0)
    desc_1, kpts_1 = model(image_1)
    
    # compute similarity matrix
    sim_mat = cosim(desc_0, desc_1)
    
    # compute the descriptor loss
    # using ground truth correspondences
    loss_desc = nll(sim_mat, corr_0, corr_1)
    
    # measure matching success
    y = is_match_success(sim_mat, corr_0, corr_1)
    
    # compute keypoint loss
    # using current matching success
    loss_kpts = bce(kpts_0, y, corr_0)
    loss_kpts += bce(kpts_1, y, corr_1)
    
    return loss_desc + loss_kpts
\end{lstlisting}
\caption{Pseudo-code: learning keypoints from a single image.}
\label{fig:algorithm}
\end{figure}

\subsection{Architecture}

\begin{figure}[h]
\caption{Architecture for \model}
\centering
\includegraphics[width=0.8\linewidth]{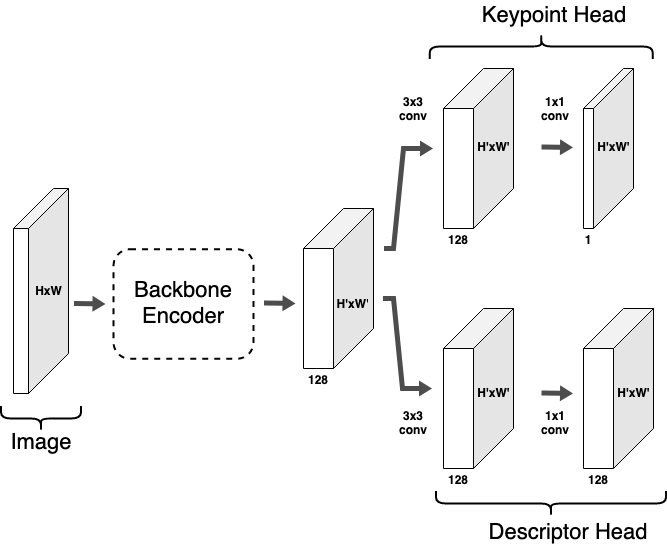}
\label{fig:arch}
\end{figure}

The \model architecture (\cref{fig:arch}) is inspired by the "detect-and-describe" architecture originally proposed by SuperPoint \cite{detone2018superpoint}. A dense feature map is first extracted by feeding an image to an encoder backbone. The shared feature map is then fed to two heads (a keypoint head and a descriptor head). 

\begin{itemize}[noitemsep]
    \item The \textit{keypoint head} extracts the logits; used to calculate the dense keypoint probabilities.
    \item The \textit{descriptor head} extracts a dense descriptor map; used later to calculate keypoint similarities.
\end{itemize}
The model is backbone-agnostic and can easily be swapped.

\subsection{High Matching Probability Defines Keypoints}

As mentioned above, the keypoint probability estimate predicts the probability that a pixel will be correctly matched (i.e. survive a round-trip). Points with the highest likelihood of matching correctly are exactly those which we select as keypoints.

A common approach \cite{detone2018superpoint,tyszkiewicz2020disk,revaud2019r2d2,dusmanu2019d2} for obtaining keypoint probabilities is to use a softmax cell-based approach. A cell is a $N\times N$ patch in which the probability of each cell position is determined by a local softmax. The softmax operates on $N\times N+1$ bins; $+1$ being the \textit{dustbin}, handling the case of cells devoid of keypoints.

\model's approach is equivalent to a cell-based approach with a cell size of $N=1$. This has several consequences. First, the softmax formulation becomes a sigmoid $\sigma(x) = \frac{1}{1+e^{-x}}$.  Second, it removes the sparsity constraint that keypoints are exclusive events inside a cell. And third, this removes a free parameter (the cell size $N$) that we do not have to later tune.

In the same spirit, \model does not use NMS during inference. Even though NMS is an established pruning technique that aims to spread out keypoints \cite{detone2018superpoint}, we find SiLK doesn't need NMS to perform (cf. \cref{tab:exp-hpatches-sparse}).



\subsection{Descriptors Define Matching Probability}



Similar to \cite{sun2021loftr,tyszkiewicz2020disk}, we model the cycle matching probability using a double softmax (i.e. the probability of matching $i$ to $j$, and back).

$$ P_{i\xleftrightarrow[]{}{}j} = P_{i\xrightarrow{}j} P_{i\xleftarrow{}j} $$
where $P_{i\xrightarrow{}j}$ is the directional probability of matching the $i$th descriptor from image $I$ to the $j$th descriptor in image $I'$. $P_{i\xleftarrow{}j}$ is similar, but in the reverse direction.  Both forward and backward probabilities are modeled as a softmax with temperature over descriptor cosine similarities.  For fixed $i$, $P_{i\xrightarrow{}j}$ is a softmax over the $i$-th row of the descriptor similarity matrix, and $P_{i\xleftarrow{}j}$ takes softmax over column $j$.

\subsection{Training}

\subsubsection{Self-Supervision}

Pixel-accurate correspondences are required during training. Similar to SuperPoint \cite{detone2018superpoint}, we obtain image-pair correspondences by applying a random transformation (homography) to an image. However, the homography is a linear mapping that gives correspondences at a subpixel level. 

To obtain pixel correspondences, we apply the sampled homography to all the pixel positions of image $I$; positions being the center of pixels (e.g. the top-left pixel has position $(0.5,0.5)$). This first step establishes dense directional correspondences from $I$ to $I'$. We then run the same process from $I'$ to $I$ (using the inverse homography this time). Once both directional correspondences are obtained, the resulting positions are discretized; out-of-bound and non-bijective correspondences are discarded.





\subsubsection{Image Augmentations}


As done in \cite{detone2018superpoint,sarlin2020superglue}, we employ image augmentation to improve robustness; augmentations include random brightness, contrast, gaussian noise, speckle noise and motion blur. We refer to supplementary materials for details.

\subsubsection{Negative and Positive Selection}


One defining property of a keypoint is \textit{distinctiveness} -- the point can be reliably distinguished from its peers.  In our case this means the point can be reliably identified in a matching algorithm, similar to \cite{truong2019glampoints,tyszkiewicz2020disk}.  With that in mind, we adopt a similar supervision technique as \cite{truong2019glampoints}. Keypoints that are correctly matched (using the currently trained descriptors) are labelled as positive; otherwise negative.  

\subsubsection{Descriptor and Keypoint Losses} \label{sec:meth-keypoint-learning}

Similar to \cite{sun2021loftr,tyszkiewicz2020disk}, the descriptor loss is the negative log-likelihood loss applied to the matching probabilities for the positive round-trips (i.e. paths from point $i$ to its location $i'$ in the transformed image and back again).
\begin{align*}
    \mathcal{L}_{desc} = -\left(\log P_{i\xrightarrow{}i'} + \log P_{i\xleftarrow{}i'}\right)
\end{align*}
This implicitly penalizes non-positive paths via softmax. One might notice $\mathcal{L}_{desc}$ requires the computation of a large matrix (size $HW\times HW$). To handle potential GPU out-of-memory, we provide a simple, yet efficient implementation which computes the similarity matrix in a block-wise fashion, and recomputes block dot products instead of storing them for backpropagation (in the same vein as \cite{rabe2021self, xFormers2022}).

The keypoint loss $\mathcal{L}_{key}$ is a simple binary cross-entropy loss applied to a logistic sigmoid, in contrast to \cite{truong2019glampoints}.
It is trained to identify keypoints with successful round-trip matches (defined by mutual-nearest-neighbor) among all others (unsuccessful).

%% file: sections/experiments.tex
\section{Experiments}

\label{sec:experiments}

In this section, we empirically evaluate \model together with representative baselines and state-of-the-art methods.  On HPatches we evaluate a suite of complementary keypoint quality metrics 
(Repeatability, Mean Matching Accuracy~\cite{1498756}) and planar stereo estimation capabilities (Homography Estimation). In addition, we benchmark on three real-world stereo tasks: outdoor camera pose estimation on Image Matching Challenge (IMC) 2022, and both indoor camera pose estimation and 3D point-cloud registration on ScanNet. In these experiments, we study the following: 



\noindent \textbf{(1)} Many methods employ complex strategies to learn and predict good keypoints and descriptors, including elements like multi-stage training, cell-based priors, complicated post-processing, context aggregation, and groundtruth 3D pose supervision (see ~\cref{tab:key-det}), in various combinations.  \textit{What machinery is necessary?} \model contains no such machinery, and can be viewed as a reduction from these methods. However, \model either achieves new SOTA or compares very favorably. This questions the need for complex schemes for the evaluated tasks. 

\noindent \textbf{(2)} We observed rather strong performance from engineered features (e.g. SIFT in \cref{tab:exp-hpatches-sparse}\&\cref{tab:exp-scannet}) vs learned methods. This motivates us to revisit design choices in a learned keypoint detector: \textit{what makes a good keypoint detector?} We ablate each model component (data, backbone, etc.) and test generalization performance under various conditions (e.g. input size, test data, task, etc). \model proves very robust to these  choices~(\cref{sec:analysis}). In particular, a very lightweight version of \model (two 3x3 convolution layers) is competitive to SOTA on homography estimation, camera pose estimation and point cloud registration (\cref{tab:ana-backbone-results}\&\cref{tab:ana-training-input-size}).






We hope these results can serve the community and help adapt keypoint models to their tasks and needs. For example, labeling tasks (e.g. self-training~\cite{10.1007/978-3-030-58565-5_3}, object pose~\cite{Reizenstein2021CommonOI}) might focus on high accuracy (i.e. larger backbone and denser keypoint selection), while tasks requiring speed (e.g. SLAM~\cite{MurArtal2015ORBSLAMAV}) might find our lightweight backbone attractive.


\subsection{Implementation Details}

\noindent Our own training pipeline has been used for all experiments with \model, as well as our reproduced SuperPoint (\cref{tab:exp-hpatches-sparse}) results. Training time is \textasciitilde 5 hours on 2 Tesla V100-SXM2 GPUs using our default setup.

\noindent \textbf{Default Setup.}
Unless specified otherwise, all results use the following setup. Trained on COCO~\cite{lin2014microsoft} images (randomly sampled), with Adam~\cite{Kingma2015AdamAM} optimizer with learning rate $1e^{-4}$ and betas $(0.9, 0.999)$; trained for 100k iterations; batch size of 1 per GPU; dense descriptor map is $146\times146$ for all architectures and input resolutions; cosine similarities scaled by temperature $20^{-1}$; \VGG-4 backbone (VGG architecture \textbf{with max-pooling removed}, details in \cref{sec:ana-backbone}); sparse keypoints obtained with top-k ($k = 10000$); detection head is 1 3x3 convolution (128-dims), and 1 1x1 convolution; descriptor head is 1 3x3 convolution (128-dims), and 1 1x1 convolution (128-dims out); no padding in convolutions; ReLU and batchnorm used as non-linearity and normalization (see supplementary).

\subsection{HPatches Homography Estimation} \label{sec:exp-hpatches}


Following \cite{detone2018superpoint,sarlin2020superglue,revaud2019r2d2,sun2021loftr}, we evaluate homography estimation on HPatches \cite{balntas2017hpatches}. HPatches contains 57 scenes (of 6 images) with significant illumination changes and 59 scenes with large viewpoint variations. Images in each scene are related by groundtruth homographies. We follow LoFTR~\cite{sun2021loftr}, (currently SOTA on HPatches), and and scale the shorter image edge to 480 at inference time.

\noindent \textbf{Evaluation Protocol.} For every pair of images, the model detects a set of keypoints.
These keypoints are desired to be distinctive and therefore repeatably detected across views. We use \textit{Repeatability} to evaluate detection performance as in~\cite{detone2018superpoint}. To test invariance of keypoint descriptors, each model's preferred matching algorithm establishes correspondence across images to obtain a subset of keypoints. We distinguish this subset as \textit{post-matching} and the entire set as \textit{pre-matching}. The accuracy of each correspondence is evaluated by \textit{Mean Matching Accuracy} as in \cite{dusmanu2019d2,revaud2019r2d2,tyszkiewicz2020disk}. Finally, we use OpenCV RANSAC algorithm to compute homography based on matching results and evaluate \textit{Homography Estimation Accuracy}~\cite{detone2018superpoint} and \textit{Homography Estimation AUC}~\cite{sarlin2020superglue,sun2021loftr}.





\noindent \textbf{Baselines.} We compare \model against both sparse detector-based methods and dense detector-free methods. The sparse detector-based methods generally follow ``detect \textit{then} match": the model first detects a sparse set of distinctive points, then matches features. We include SIFT~\cite{lowe2004distinctive} as well as learned detectors SuperPoint~\cite{detone2018superpoint} (both official release and our repro), R2D2~\cite{revaud2019r2d2} and DISK~\cite{tyszkiewicz2020disk}. On the other hand, dense detector-free perform ``detection \textit{by} matching": the model first extracts features, then applies a learned pre-matching CA module (e.g. GNN~\cite{sarlin2020superglue} or transformer~\cite{sun2021loftr}) to adapt the features to a specific pair of images, and then finds matches. While \model does not employ CA, we still include comparisons to the SOTA detector-free LoFTR~\cite{sun2021loftr}.


\begin{table*}
  \centering
  \footnotesize
  \begin{tabular}{l|cc|cc|cc|cc|cc}
 \hline
 & \multicolumn{2}{c|}{Repeatability} & \multicolumn{2}{c|}{Hom. Est. Acc.} & \multicolumn{2}{c|}{Hom. Est. AUC} & \multicolumn{2}{c|}{MMA} & \multicolumn{2}{c}{\# of keypoints} \\
 \hline
 & \footnotesize{$\epsilon=1$} & \footnotesize{$\epsilon=3$} & \footnotesize{$\epsilon=1$} & \footnotesize{$\epsilon=3$} & \footnotesize{$\epsilon=1$} & \footnotesize{$\epsilon=3$} & \footnotesize{$\epsilon=1$} & \footnotesize{$\epsilon=3$} & \footnotesize{pre-match} & \footnotesize{post-match} \\
 \hline
 SuperPoint (MagicLeap) & 0.34 & 0.61 & 0.43 & 0.8 & 0.2 & 0.51 & 0.41 & $0.72$ & 847 & 499 \\
 SuperPoint (Ours) & 0.33 & 0.52 & 0.48 & 0.75 & 0.26 & 0.52 & 0.38 & 0.53 & 1143 & 474 \\
 SIFT & 0.31 & 0.52 & 0.6 & 0.84 & 0.34 & 0.61 & 0.41 & 0.55 & 2189 & 910 \\ 

 R2D2 & 0.36 & 0.72 & 0.45 & 0.79 & 0.2 & 0.5 & 0.34 & 0.75 & 6088 & 1967 \\
 DISK & 0.38 & 0.69 & 0.45 & 0.8 & 0.22 & 0.52 & 0.52 & \textbf{0.84} & 3349 & 1794 \\ 
 \hline
 \hline
 \model (top-10k) & \textbf{0.62} & \textbf{0.81} & \textbf{0.62} & \textbf{0.87} & \textbf{0.4} & \textbf{0.66} & \textbf{0.59} & 0.71 & 10000 & 4283 \\
 \model (top-5k) & 0.56 & 0.76 & 0.6 & 0.85 & 0.39 & 0.64 & 0.57 & 0.69 & 5000 & 2074 \\
 \model (top-1k) & 0.43 & 0.61 & 0.53 & 0.81 & 0.32 & 0.58 & 0.52 & 0.63 & 1000 & 389 \\
 \hline
 \end{tabular}
 \caption{\textbf{\model achieves new SOTA on HPatches compared to other methods with sparse keypoints and features.} Despite its simplicity, \model achieves higher performance on all metrics except MMA@3. We include the \# of keypoints to ensure a fair comparison.}
 \label{tab:exp-hpatches-sparse}
\end{table*}

\begin{table}
\setlength{\tabcolsep}{3.5pt}
  \centering
  \footnotesize
  \begin{tabular}{l|cc|cc|cc}
 \hline
 & \multicolumn{2}{c|}{Hom. Est. Acc.} & \multicolumn{2}{c|}{Hom. Est. AUC} & \multicolumn{2}{c}{MMA} \\
 & \footnotesize{$\epsilon=1$} & \footnotesize{$\epsilon=3$} & \footnotesize{$\epsilon=1$} & \footnotesize{$\epsilon=3$} & \footnotesize{$\epsilon=1$} & \footnotesize{$\epsilon=3$} \\ 
 \hline
 LoFTR (MegaDepth) & \textbf{0.65} & \textbf{0.87} & 0.37 & 0.65 & \textbf{0.64} & \textbf{0.91}  \\
 LoFTR (ScanNet) & 0.24 & 0.57 & 0.07 & 0.33 & 0.36 & 0.76  \\
 \hline
 \model (top-10k) & 0.62 & \textbf{0.87} & \textbf{0.4} & \textbf{0.66} & 0.59 & 0.71  \\
 \model (top-5k) & 0.60 & 0.85 & 0.39 & 0.64 & 0.57 & 0.69  \\
 \hline
 \end{tabular}
 \caption{\textbf{\model achieves competitive performance to SOTA LoFTR on HPatches, despite not using context aggregation.} The top-5k \model has similar number of matches compared to outdoor LoFTR. We remark that LoFTR has large generalization gap when training on different types of dataset (indoor vs. outdoor), possibly due to the contexualizer.}
 \label{tab:exp-hpatches-dense}
\end{table}

\noindent \textbf{Results.} Despite its simplicity, \model outperforms all methods on repeatability, homography accuracy and homography estimation (\cref{tab:exp-hpatches-sparse}). In particular, \model has a strong margin when the error threshold is small ($\epsilon=1$). This validates our pixel-accurate keypoint localization.  \model lags only on the MMA@3 metric. This may be caused by our pixel accurate contrastive loss, which amplifies local descriptor differences and makes local matching errors less likely.  Consequently, \model does not benefit from increasing error threshold in MMA. In addition, even vs LoFTR (which uses dense features and CA) in~\cref{tab:exp-hpatches-dense}, \model shows strong performance on Homography Estimation AUC and competitive performance on Homography accuracy.  This questions the necessity of CA for these particular tasks; this may be valuable in applications which are particularly sensitive to runtime performance.

\subsection{IMC 2022 outdoor pose estimation} \label{sec:IMC_exp}


The Image Matching Challenge (\href{https://www.kaggle.com/competitions/image-matching-challenge-2022/overview/description}{IMC 2022})~\cite{Jin2020IMC} provides pairs of outdoor images from different viewpoints; participants are required to estimate the fundamental matrix.
Camera pose accuracy is then computed for ten thresholds of rotation and translation error (ranging from $(1^{\circ}, 20\mathrm{cm})$ to $(10^{\circ}, 5\mathrm{m})$). Mean average accuracy (mAA) is reported by averaging across thresholds and scenes.




\textbf{Evaluation Protocol.} Sparse methods, such as DISK and SiLK, detect keypoints in individual images; mutual nearest neighbor is used to select matches from each pair. Methods with CA, such as SuperGlue and LoFTR, directly identify matches from each pair of images. In either case, results are passed to MAGSAC~\cite{barath2019magsac} to estimate camera pose. The challenge allows different image sizes and tuning MAGSAC parameters~\cite{Jin2020IMC}.  We use 30k keypoints and MAGSAC threshold $.25$.


\textbf{Baselines.} We consider the best leaderboard results from three baselines. (i) DISK is the winner of IMC 2020 and is SOTA among sparse methods. 2022 DISK results are from the IMC team (\href{https://www.kaggle.com/code/eduardtrulls/imc2022-baseline-submission-disk}{submission}). We also take the best version of (ii) SuperGlue (\href{https://www.kaggle.com/code/yufei12/superglue-baseline}{submission}) and (iii) LoFTR (\href{https://www.kaggle.com/code/ammarali32/imc-2022-kornia-loftr-from-0-533-to-0-721}{submission}) provided by the community.

\textbf{Results.} We coarsely tune \model for this task for 30 trials (compared to LoFTR's 200 trials). \model again performs competitively (c.f. \cref{tab:result-imc}), outperforming DISK by a significant margin (\textbf{+0.19/+0.18} mAA). \model also performs favorably compared to SuperGlue, which uses context aggregation and optimal transport matching. 


\begin{table}
\setlength{\tabcolsep}{3.5pt}
  \centering
  \footnotesize
  \begin{tabular}{ l|c|c  }
 \hline
 & private mAA & public mAA \\
 \hline
 \multicolumn{3}{l}{\textbf{Sparse Features}} \\
 DISK & 0.491 & 0.502 \\
 SuperGlue & 0.676 & 0.678 \\
 SiLK & \textbf{0.685} & \textbf{0.684} \\
 \hline
 \multicolumn{3}{l}{\textbf{Dense Features}} \\ 
 LoFTR (MegaDepth) & \textbf{0.735} & \textbf{0.721} \\
 \hline
 \end{tabular}
 \caption{\textbf{\model achieves new SOTA for sparse methods on IMC2022 and performs competitively to dense methods with context aggregation.}}
 \label{tab:result-imc}
\end{table}

\subsection{ScanNet: Indoor Pose \& Point Clouds} \label{sec:scannet_exp_all}




ScanNet~\cite{dai2017scannet} is a large-scale dataset of 1513 indoor scenes of RGB-D images and ground-truth camera poses. Using the official train/val/test split we evaluate both relative camera pose estimation and point-cloud registration. Relative camera pose estimation has been used in multiple previous works~\cite{Yi2017LearningTF,zhang2019oanet,9008398,sarlin2020superglue,sun2021loftr}. The task is to estimate the essential matrix with RANSAC from point matches in a pair of images. We report pose error AUC at thresholds ($5^\circ$,$10^\circ$,$20^\circ$) using 20k keypoints and RANSAC threshold $.5$.

This protocol measures translation error in degrees and is known to suffer from scale ambiguity~\cite{Yi2017LearningTF}. Angular translation error may be unstable, in particular if the underlying translation (in meters) is small. In response, recent works~\cite{9157740,el2021unsupervisedr} have introduced a 3D point-cloud registration task, using the ground-truth depth provided in ScanNet. A pair of images 20 frames apart is first sampled.  Given this pair, a model predicts point matches. After matching, relative camera pose is estimated. Different from the previous protocol, ground-truth depth and camera intrinsics are now used to align matches in 3D. In addition to relative pose errors (reported separately for translation (cm) and rotation (degree)), the Chamfer distance (in cm) is measured between the registered point cloud and the groundtruth point cloud. We refer to the original papers~\cite{9157740,el2021unsupervisedr} for details on the evaluation setup and metrics.





\subsubsection{Relative pose estimation}\label{sec:scannet_old}

\noindent \textbf{Baselines.} We compare \model with both sparse detector-based methods and dense detector-free methods. For the sparse methods, we consider local feature descriptors (R2D2, SuperPoint) with mutual nearest neighbor (MNN) for matching. \model falls into this category. In addition, we consider multiple learned context aggregation methods for matching that operates on SuperPoint, including PointCN~\cite{Yi2017LearningTF}, OANet~\cite{zhang2019oanet} and SuperGlue. For detector-free dense methods, we consider DRC-Net~\cite{li20dualrc} and LoFTR. We include two versions of LoFTR: one trained on MegaDepth with optimal transport post-processing and one trained on ScanNet.

\noindent \textbf{Results.} As summarized in \cref{tab:result-scannet-old}, \model significantly outperforms D2-Net (\textbf{+12.7}) and SuperPoint  (\textbf{+8.6}) when using mutual nearest neighbor matching. In addition, \model outperforms the previous SOTA sparse method SuperGlue, despite its simpler design without context aggregation. \model performs similarly to LoFTR trained with MegaDepth. \model is only outperformed by LoFTR trained on ScanNet, the same as evaluation data. 

\begin{table}
\setlength{\tabcolsep}{3.5pt}
  \centering
  \footnotesize
  \begin{tabular}{ l|c|c|c  }
 \hline
 Pose Estimation AUC & @$5^\circ$ & @$10^\circ$ & @$20^\circ$\\
 \hline
 \multicolumn{2}{l}{\textbf{Sparse Features}} \\
 D2-Net~\cite{dusmanu2019d2} + MNN & 5.3 & 14.5 & 28.0 \\
 SuperPoint~\cite{detone2018superpoint} + MNN & 9.4 & 21.5 & 36.4 \\
 SuperPoint + PointCN~\cite{Yi2017LearningTF} & 11.4 & 25.5 & 41.4 \\
 SuperPoint + OANet~\cite{zhang2019oanet} & 11.8 & 26.9 & 43.9 \\
 SuperPoint + SuperGlue~\cite{sarlin2020superglue} & 16.2 & \underline{33.8} & \textbf{51.8} \\
 SiLK + MNN & \textbf{18.0} & \textbf{34.4} & \underline{50.4}\\
 \hline
 \multicolumn{2}{l}{\textbf{Dense Features}} \\
 DRC-Net~\cite{li20dualrc} & 7.7 & 17.9 & 30.5 \\
 LoFTR (MegaDepth) & 16.9 & 33.6 & 50.6 \\
 LoFTR (ScanNet) & \textbf{21.5} & 40.8 & 57.6 \\
 \hline
 \end{tabular}
 \caption{\textbf{\model advances SOTA on relative pose estimation among sparse methods on ScanNet and performs competitively against dense method LoFTR.}}
 \label{tab:result-scannet-old}
\end{table}

\subsubsection{Pairwise 3D point-cloud registration}

\noindent \textbf{Baselines.} We consider three main types of baselines. \textbf{(i)} \textbf{Sparse Features + RANSAC} We extract sparse keypoints and their features from off-the-shelf models, and use RANSAC to estimate alignment. This includes SIFT and SuperPoint, and 3D geometry model FCGF~\cite{9009829}. \textbf{(ii)} \textbf{Dense Feature Matching} We follow the \cite{el2021unsupervisedr} and select high-quality corresponding pairs using the ratio test, and then solve a weighted Procustes problem~\cite{choy2020deep,Kabsch:a12999} to produce alignment. We add a dense version of SuperPoint by discarding the keypoint prediction. We include the current state-of-the-art URR~\cite{el2021unsupervisedr} that learns invariant point descriptors through cross-view synthesis. By ignoring the keypoint scoring prediction, \model also belongs to this category. The goal is to evaluate the quality of the dense point features. \textbf{(iii)} \textbf{Pose/Geometry Supervised} We consider methods that use groundtruth poses to supervise, which are not required in (i) and (ii). These include LoFTR, DGR~\cite{choy2020deep} and 3D MV Registration~\cite{9157740}. For LoFTR, we include only the MegaDepth model, since it performed better in this task.

\noindent \textbf{Results.} As shown in \cref{tab:exp-scannet}, \model achieves new SOTA across all metrics (except rotation accuracy at $45^{\circ}$). In particular, \model achieves very high accuracy at small thresholds ($5^{\circ}$ angular, 5cm translation and 1cm for Chamfer), validating \model's pixel-level precision. Comparing with DGR, 3D MV Reg and LoFTR that use groundtruth camera poses during training, \model significantly outperforms, indicating that groundtruth 3D supervision is not necessary to train good keypoint features. We did not include the Chamfer results for LoFTR as the provided positions do not match the required resolution for correct Chamfer evaluation. \model also achieves superior performance vs previous SOTA URR. We note that URR requires two different frames sampled from the same scene during training, and is supervised by an advanced differentiable cross-view rendering process. \model, on the other hand, only requires a single 2D image and is trained with a simple point matching loss. Finally, we observe that SuperPoint performs competitively when evaluated in this dense fashion; this is an important difference from the results reported in URR using sparse features.

\begin{table*}
\setlength{\tabcolsep}{3.5pt}
  \centering
  \footnotesize
  \begin{tabular}{l|ccc|cc|ccc|cc|ccc|cc}
 \hline
  & \multicolumn{5}{c|}{Rotation} & \multicolumn{5}{c|}{Translation} & \multicolumn{5}{c}{Chamfer} \\
 \hline
  & \multicolumn{3}{c|}{Accuracy $\uparrow$} & \multicolumn{2}{c|}{Error $\downarrow$} & \multicolumn{3}{c|}{Accuracy $\uparrow$} & \multicolumn{2}{c|}{Error $\downarrow$} & \multicolumn{3}{c|}{Accuracy $\uparrow$} & \multicolumn{2}{c}{Error $\downarrow$} \\
 \hline
 &  $5^{\circ}$ & $10^{\circ}$ & $45^{\circ}$ & Mean & Med. & 5 & 10 & 25 & Mean & Med. & 1 & $5$ & 10 & Mean & Med. \\
 \hline
 \multicolumn{15}{l}{\textbf{Sparse Features + RANSAC}} \\
 SIFT~\cite{lowe2004distinctive} & 55.2 & 75.7 & 89.2 & 18.6 & 4.3 & 17.7 & 44.5 & 79.8 & 26.5 & 11.2 & 38.1 & 70.6 & 78.3 & 42.6 & 1.7 \\
 SuperPoint~\cite{detone2018superpoint} & 65.5 & 86.9 & 96.6 & 8.9 & 3.6 & 21.2 & 51.7 & 88.0 & 16.1 & 9.7 & 45.7 & 81.1 & 88.2 & 19.2 & 1.2 \\
 FCGF~\cite{9009829} & 70.2 & 87.7 & 96.2 & 9.5 & 3.3 & 27.5 & 58.3 & 82.9 & 23.6 & 8.3 & 52.0 & 78.0 & 83.7 & 24.4 & 0.9 \\

 \hline
 \multicolumn{15}{l}{\textbf{Pose/Geometry Supervised}} \\
 DGR~\cite{choy2020deep} & 81.1 & 89.3 & 94.8 & 9.4 & 1.8 & 54.5 & 76.2 & 88.7 & 18.4 & 4.5 & 70.5 & 85.5 & 89.0 & 13.7 & 0.4 \\
 3D MV Reg~\cite{9157740} & 87.7 & 93.2 & 97.0 & 6.0 & 1.2 & 69.0 & 83.1 & 91.8 & 11.7 & 2.9 & 78.9 & 89.2 & 91.8 & 10.2 & 0.2 \\
 LoFTR(MegaDepth)~\cite{sun2021loftr} & 91.7 & 96.8 & 99.4 &  2.8 &  1.2 & 65.9 & 85.2 & 97.1 &  6.0 &  3.3 &  - & - & - & - & - \\
 
 \hline
 \multicolumn{15}{l}{\textbf{Dense feature matching}} \\
SuperPoint~\cite{detone2018superpoint} & 93.0 & 98.4 & \textbf{99.8} &  2.5 &  1.6 & 56.8 & 84.7 & 98.2 &  6.5 &  4.3 & 77.3 & 96.1 & 98.4 &  4.5 &  0.3 \\
URR~\cite{el2021unsupervisedr} & 92.7 & 95.8 & 98.5 & 3.4 & \textbf{0.8} & 77.2 & 89.6 & 96.1 & 7.3 & 2.3 & 86.0 & 94.6 & 96.1 & 5.9 & \textbf{0.1} \\
 \model(\VGG-4) & \textbf{98.1} & \textbf{99.0} & 99.6 &  \textbf{1.7} & \textbf{0.8} & \textbf{82.9} & \textbf{94.8} & \textbf{99.0} & \textbf{4.1} &  \textbf{2.1} & \textbf{92.8} & \textbf{98.3} & \textbf{99.1} &  \textbf{4.3} &  \textbf{0.1} \\
 \hline
 \end{tabular}
 \caption{\textbf{\model achieves state-of-the-art on camera pose estimation and point cloud registration on ScanNet.}}
 \label{tab:exp-scannet}
\end{table*}

%% file: sections/analysis.tex
\subsection{What makes good keypoint detectors?} \label{sec:analysis}


Leveraging \model's flexibility (\cref{sec:methodology}), we comprehensively ablate a large pool of design choices such as model architecture and image resolution. Surprisingly, we found that reducing architecture size, compute cost, and training input size only mildly impact model performance on homography estimation, camera pose estimation and point cloud registration. This benefits many important applications, such as on-device inference.

Here we discuss the key findings. We use the metrics Repeatability@1 (R), Homography Estimation Accuracy@1 (HA), Mean Matching Accuracy@1 (MMA) for HPatches, and Rotation Accuracy@$5^{\circ}$ (RA), Translation Accuracy@5cm (TA) and Chamfer@1 (C) for ScanNet, all at lowest error thresholds. Additional results and analysis are included in supplementary.


\subsubsection{Agnostic to backbone}
\label{sec:ana-backbone}


Existing methods use various backbones (as shown in \cref{tab:model-learning}); the effects on keypoint models are not well understood. We consider FPN from LoFTR and UNet from DISK; both are modern compared to \model's \VGG~\cite{Simonyan2015VeryDC} backbone.  We find no empirical performance gain despite far greater parameter counts (\cref{tab:ana-backbone-results}). This questions the need for high-capacity models for these keypoint problems.

Next we reduce the complexity of the original SuperPoint backbone VGG-4. \textbf{Max-pooling and up-sampling layers are removed}. Our \VGG-4 contains four convolution blocks, each with two convolution layers followed by ReLU. We discard convolution blocks from \VGG-4 to obtain \VGG-3, \VGG-2 and \VGG-1. On top of \VGG-1, we reduce channel and descriptor sizes to 64 and 32 respectively and obtain an ultra-lightweight model, \VGG-\textmu.  Results on repeatability, homography estimatimation, camera pose estimation, and point cloud registration only drop mildly as we shrink the model. In particular, \model(\VGG-\textmu\space in \cref{tab:ana-backbone-results}) achieves very competitive performance vs SOTA (\cref{tab:exp-hpatches-sparse}\&\cref{tab:exp-scannet}).  On the other hand, keypoint matching (MMA) scores drop signficantly.  We suggest two possible reasons:  first, pointwise matching may benefit from the larger receptive field of deeper models.  Second, homography estimation aggregates numerous pointwise measurements; homographies will improve if the noises cancel out, or if the worst estimates are removed by RANSAC.






\begin{table}
\setlength{\tabcolsep}{2.5pt}
  \centering
  \footnotesize
  \begin{tabular}{ l|ccc|ccc|ccc  }
 \hline
 & \multicolumn{3}{c|}{HPatches}  & \multicolumn{3}{c|}{ScanNet} & \multicolumn{3}{c}{Model} \\
 & R & HAc & MMA & RA & TA & C & Param & FPS & GFLOP \\
 \hline
 \VGG-4 & 0.62 & \textbf{0.62} & \textbf{0.59} & \textbf{98.1} & 82.9 & 92.8 & 942k & 12.2 & 370\\
 \VGG-3 & 0.63 & 0.61 & 0.58 & 98.0 & \textbf{84.2} & \textbf{93.5} & 868k & 12.5 & 330\\
 \VGG-2 & 0.63 & 0.57 & 0.55 & 97.3 & 83.5 & 92.4 & 757k & 14.3 & 268\\
 \VGG-1 & 0.63 & 0.57 & 0.44 & 94.6 & 82.1 & 90.0 & 461k & 18.9 & 90\\
 \VGG-\textmu & \textbf{0.64} & 0.56 & 0.40 & 93.5 & 81.1 & 89.0 & \textbf{76k} & \textbf{36.5} & \textbf{23}\\
 \hline \hline
 FPN~\cite{sun2021loftr} & 0.58 & 0.55 & 0.52 & - & - & - & 6.6M & 17.5 & 298 \\
 \hline
 UNet~\cite{tyszkiewicz2020disk} & 0.60 & 0.41 & 0.58 & - & - & - & 1.3M & 25.9 & 198 \\
 \hline
 \end{tabular}
 \caption{\textbf{\model is backbone agnostic.} Backbones from existing methods are trained and evaluated. Low-capacity model perform well on Repeatability, Homography Estimation Accuracy, Camera Pose Estimation and Point Cloud Registration, but drops on Mean Matching Accuracy. FPS and GFLOPs measured on 480$\times$640 images with NVIDIA Quadro GP100 GPU. 
 }
 \label{tab:ana-backbone-results}
\end{table}

\subsubsection{Fast training on tiny images}

By default, \model uses 146x146 descriptor feature map resolution during training.
In areas like object detection or image recognition, higher resolution typically leads to stronger results. In \model, higher resolution provides more points. This benefits the contrastive loss with more negatives, but also increases training time and GPU memory usage.

Surprisingly, performance changes very little when varying resolution during training (\cref{tab:ana-training-input-size}), especially on ScanNet. Tiny feature maps (82x82) remain competitive on both HPatches and ScanNet, and trains in 1.7 hours on two GPUs. This enables applications like test-time finetuning, on-device finetuning, and rapid experimental iteration.

%



\begin{table}
\setlength{\tabcolsep}{3.5pt}
  \centering
  \footnotesize
  \begin{tabular}{c|c|ccc|ccc}
 \hline
 \multicolumn{2}{c|}{} & \multicolumn{3}{c|}{HPatches} & \multicolumn{3}{c}{ScanNet} \\
 Size & Time & R & HAc & MMA & RA & TA & C \\
 \hline
 $82^2$ & \textbf{1.7h} & 0.60 & 0.58 & 0.56 & 98.1 & \textbf{83.5} & 92.5 \\
 $114^2$ & 2.7h & 0.62 & \textbf{0.62} & 0.59 & 98.1 & 82.9 & 92.8 \\
 $146^2$ & 5h & \textbf{0.63} & 0.59 & 0.59 & \textbf{98.2} & 83.3 & \textbf{92.9} \\
 $178^2$ & 9.5h & \textbf{0.63} & \textbf{0.62} & 0.59 & 98.1 & \textbf{83.5} & \textbf{92.9} \\
 $210^2$ & 18h & \textbf{0.63} & 0.61 & \textbf{0.60} & 98.1 & 83.4 & 92.8 \\
 \hline
 \end{tabular}
 \caption{\textbf{Decreasing training image size has minimal impact.} This suggest \model can be trained under 3h, with little performance drop.}
 \label{tab:ana-training-input-size}
\end{table}

\subsubsection{Robustness to training data}

Existing methods use various training sets (\cref{tab:model-learning}); empirically we observe cases of poor generalization across datasets. For example, LoFTR\cite{sun2021loftr}, trained on indoor ScanNet data, drops significantly vs LoFTR trained on outdoor MegaDepth data (\cref{tab:exp-hpatches-dense}) and vice-versa (\cref{tab:result-scannet-old}).  This overfitting may be exacerbated by the high-capacity machinery used by these methods, e.g. LoFTR's Transformer contexualizer. We measure \model's robustness on training data choices, by using different images from COCO\cite{lin2014microsoft}, ImageNet\cite{deng2009imagenet}, MegaDepth\cite{li2018megadepth} and ScanNet\cite{dai2017scannet}. We also combine them to formulate a diversified set of training data. 

\model is quite robust to change in training set, with the exception of ScanNet (\cref{tab:ana-dataset}). We hypothesize this is due to the significant amount of uniform surfaces (e.g. walls, doors) present in ScanNet. These featureless areas contain few keypoints to learn from. \model's drop agrees directionally with LoFTR's drop observed in \cref{tab:exp-hpatches-dense}, but the magnitude is much smaller.  This may be because \model's \VGG\  backbone has much lower capacity than LoFTR (FPN+Transformer), and hence is less susceptible to overfitting. Finally, we remark that \model trained with COCO is used for comparisons in \cref{sec:exp-hpatches}, \cref{sec:IMC_exp} and \cref{sec:scannet_exp_all} for HPatches, IMC and ScanNet, whereas LoFTR requires different training data (MegaDepth or ScanNet) to achieve strong performance.

\begin{table}
\setlength{\tabcolsep}{3.5pt}
  \centering
  \footnotesize
  \begin{tabular}{ l|ccc|ccc  }
 \hline
 & \multicolumn{3}{c|}{HPatches} & \multicolumn{3}{c}{ScanNet} \\
 & R & HAc & MMA & RA & TA & C \\
 \hline
 COCO & 0.62 & \textbf{0.62} & \textbf{0.59} & \textbf{98.1} & 82.9 & 92.8 \\
 ImageNet & 0.63 & 0.6 & \textbf{0.59} & \textbf{98.1} & \textbf{83.5} & \textbf{93.0} \\
 MegaDep. & 0.62 & 0.61 & 0.57 & 97.9 & \textbf{83.5} & 92.9 \\
 ScanNet &  0.60 & 0.55 & 0.48 & 97.6 & 82.8 & 92.7 \\
 \hline \hline
 \textbf{C+I+M+S} & 0.61 & 0.59 & 0.54 & 97.7 & 83.0 & 92.6 \\
 \textbf{C+I+M} & \textbf{0.64} & 0.6 & \textbf{0.59} & 98.0 & 82.9 & 92.8 \\
 \hline
 \end{tabular}
 \caption{ \textbf{\model is robust to different training sets.} A noticeable drop is observed only when training on ScanNet.}
 \label{tab:ana-dataset}
\end{table}

%% file: sections/conclusion.tex
\section{Conclusion}
\label{sec:conclusion}

This paper presents \model, a simple and flexible framework for keypoint detection and descriptors. \model is designed from the principles of distinctiveness and invariance, and achieves or advances SOTA on key low-level tasks for 3D visual perception. \model's simplicity questions the need for complex machinery for good keypoint detection in low-level applications. In addition, extensive ablations reveal \model's robustness to backbone, training data and training input size. These findings lead to a tiny version of \model that is lightweight, accurate, and trains quickly. We view this ``tiny and learned" regime as very promising for applications where runtime and/or power consumption is critical. We hope \model can draw attention to the field and facilitate stronger solutions.

%% file: suppl/math.tex
\section{Maths}

In this section, we provide a short mathematical description of our method (probabilistic modeling and losses).

\subsection{Definitions}

\begin{itemize}
  \small
  \item $I$, $I'$ is the pair of images to match.
  \item $q_i$, $q'_j$ $\in\mathbb{R}$ are the $i$th and $j$th keypoint logits from $I$ and $I'$ respectively ($M$ in total).
  \item $d_i$, $d'_j$ $\in \mathbb{R}^{128}$ are the $i$th and $j$th descriptors from $I$ and $I'$ respectively ($M$ in total).
  \item $s_{ij} \in [-1,+1]$ is the similarity score between keypoint $i$ from $I$ and keypoint $j$ from $I'$. $s$ is therefore a $M\times M$ matrix.
  \item $c_i$, $c'_j$ are the $i$th and $j$th correspondence indices from $I$ and $I'$ respectively ($N$ in total, with $N \leq M$). Such that $c_i$ and $c'_i$ represent the correspondence between descriptor $d_{c_i}$ and $d'_{c'_i}$ with similarity $s_{c_ic'_i}$.
\end{itemize}


\subsection{Matching Probabilities}

\begin{figure}[h]
\centering
\includegraphics[width=0.8\linewidth]{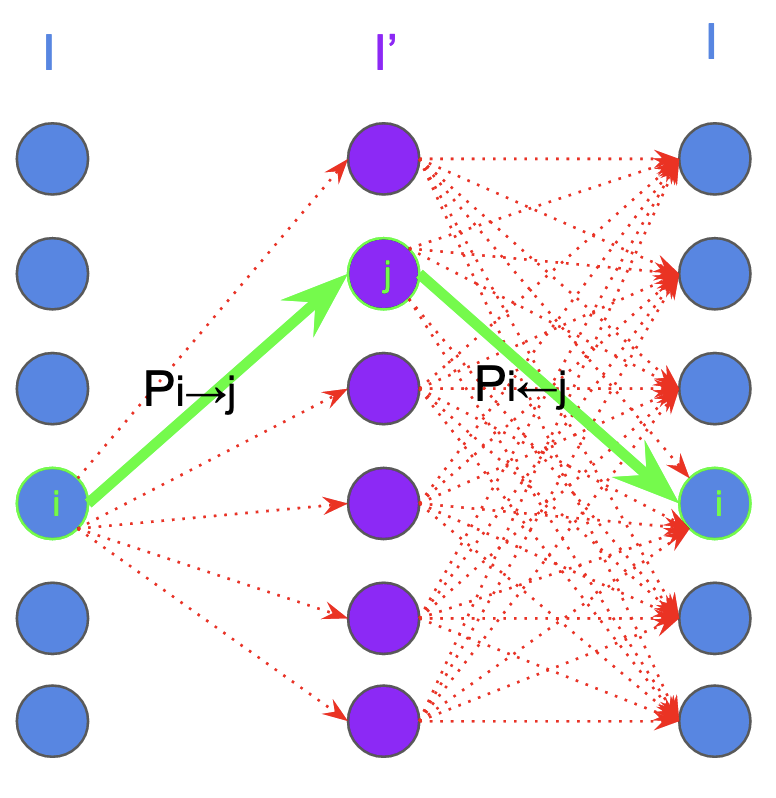}
\caption{Visualization of the cycle-consistent probabilistic path. The probability $P_{i\xleftrightarrow[]{}{}j}$ is the probability of following the green path, over the set of all possible red paths; from image $I$ to $I'$ and back.}
\label{fig:cycle-consistency}
\end{figure}

The matching probabilities are modeled by a double-softmax, enforcing the cycle-consistency property (cf. \cref{fig:cycle-consistency}).

$$ P_{i\xleftrightarrow[]{}{}j} = P_{i\xrightarrow{}j} P_{i\xleftarrow{}j} $$

\begin{itemize}
    \item $ P_{i\xrightarrow{}j} = \frac{e^{\frac{s_{ij}}{\tau}}}{\sum\limits_k{e^{\frac{s_{ik}}{\tau}}}} $ is the directional probability of matching $d_i$ to $d'_j$
    \item $ P_{i\xleftarrow{}j} = \frac{e^{\frac{s_{ij}}{\tau}}}{\sum\limits_k{e^{\frac{s_{kj}}{\tau}}}} $ is the directional probability of matching $d'_j$ to $d_i$
\end{itemize}
where $\tau$ is the temperature, and $s$ the pairwise similarity matrix; obtained using a standard cosine similarity function.
    $$s_{ij} = \mathrm{cosim}(d_i, d'_j) = \frac{\langle d_i, d'_j \rangle}{\sqrt{\langle d_i, d_i\rangle\langle d'_j, d'_j\rangle}}$$

where $\langle ., .\rangle$ is the dot product.

\subsection{Keypoint Probabilities}

All keypoints probabilities are obtained using a simple sigmoid.

$$\sigma(q_i) = \frac{1}{1+e^{-q_i}}$$

\subsection{Matching Loss}

The matching loss of a single image pair is the negative log likelihood loss, summed over the entire set of correspondences.

\begin{align*}
    \mathcal{L}_{desc} &= \mathrm{NLL}(s, c, c') \\
    &= - \frac{1}{N} \sum\limits_{i=0}^{N-1} \log P_{c_i\xleftrightarrow[]{}{}c'_i} \\
    &= - \frac{1}{N} \sum\limits_{i=0}^{N-1} \big[ \log P_{c_i\xrightarrow{}c'_i} + \log P_{c'_i\xleftarrow{}c_i} \big]
\end{align*}

\subsection{Keypoint Loss}

Once the matching success of descriptors as been measured (and stored in variable $y \in \{0,1\}^N$, cf. \cref{sec:match-success}), we can learn the keypoint probabilities using a standard binary cross-entropy loss; using the keypoint logits extracted from both $I$ and $I'$.

\begin{align*}
    \mathcal{L}_{key} = \mathrm{BCE}(q, y, c) + \mathrm{BCE}(q', y, c')
\end{align*}

where 

{\small
\begin{align*}
    \mathrm{BCE}(q, y, c) = - \frac{1}{N}\sum\limits_{i=0}^{N-1} \bigg[y_i \log \sigma(+q_{c_i}) + (1-y_i) \log \sigma(-q_{c_i}) \bigg]
\end{align*}
}

\subsection{Matching Success} \label{sec:match-success}

The matching success is the process of measuring whether or not matching currently learned descriptors (using a simple mutual nearest neighbor) would produce correct matches. It can be expressed mathematically as verifying whether or not the similarity of a ground truth correspondence is the row and column maximum in $s$.

\begin{align*}
    y_i = \mathbf{1}\big[s_{c_ic'_i} \ge \max\limits_{k}\{s_{c_ik}\}\big] \mathbf{1}\big[s_{c_ic'_i} \ge \max\limits_{k}\{s_{kc'_i}\}\big]
\end{align*}

where $\mathbf{1}\big[.\big]$ is the indicator function.

%% file: suppl/experiments.tex
\section{Additional Experiments}

In this section, we present additional experiments as evidence of the robustness of \model under varying, but realistic, conditions. We also hope this comprehensive set of data points can be used by the community to tune their own version of \model to a specific use case or task.
For example, if a task is sensitive to false positive matching, one might consider using the ratio-test filtering, as indicated in \cref{tab:exp-dist-pp}.

\model's robustness is tested against three important types of variations, as we aim to answer the following.

\textit{1) Are \model's results robust across different image resolutions ?} (cf. \cref{tab:exp-ana-inference-size,fig:ana-inference-input-size})

\textit{2) Can we improve SiLK by simply selecting more keypoints
? Or do we reach saturation for a certain value of k ?} (cf. \cref{tab:exp-top-k,fig:plot-top-k})

\textit{3) Do existing false-positive removal techniques work on \model ? And how is performance affected by it ?}(cf. \cref{tab:exp-dist-pp}).


Additionally, we empirically demonstrate the importance of \textbf{not} using zero-padding when learning keypoints (cf. \cref{tab:exp-padding}). This is an often under-emphasized point we shed light on here. 

\input{suppl/experiments/size.tex}
\input{suppl/experiments/topk.tex}
\input{suppl/experiments/dist-post-proc.tex}
\input{suppl/experiments/padding.tex}

%% file: suppl/experiments/size.tex
\subsection{Downsizing images is better than increasing $\epsilon$}

\begin{figure*}[h]
\centering
\begin{subfigure}{0.48\linewidth}
  \centering
  \includegraphics[width=\linewidth]{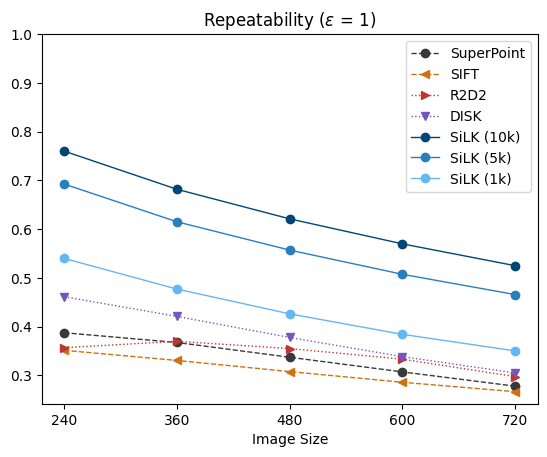}
  \label{fig:plot-r1}
\end{subfigure}
\begin{subfigure}{0.48\linewidth}
  \centering
  \includegraphics[width=\linewidth]{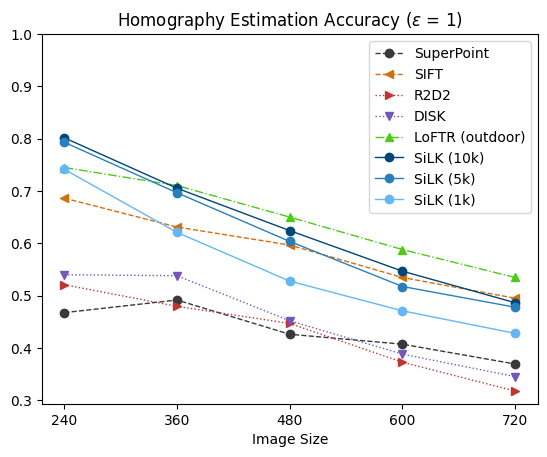}
  \label{fig:plot-hac1}
\end{subfigure}
\begin{subfigure}{0.48\linewidth}
  \centering
  \includegraphics[width=\linewidth]{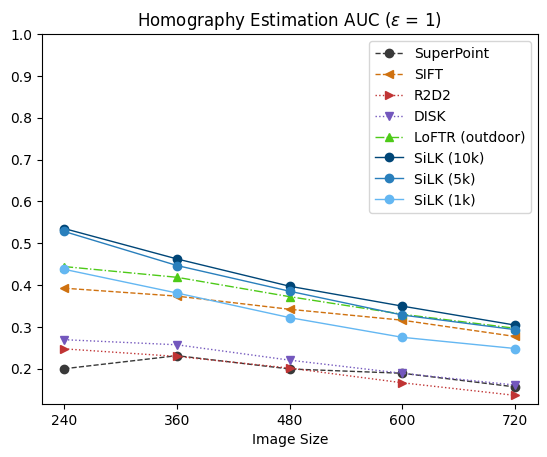}
  \label{fig:plot-hau1}
\end{subfigure}
\begin{subfigure}{0.48\linewidth}
  \centering
  \includegraphics[width=\linewidth]{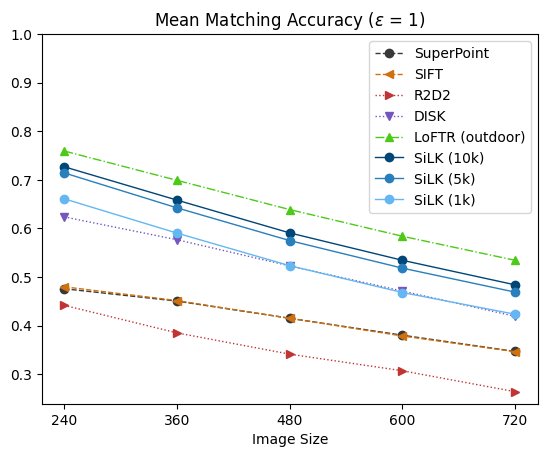}
  \label{fig:plot-mma1}
\end{subfigure}
  \caption{\textbf{\model's performance is robust across different input scales.} All sparse methods are outperformed by \model on all metrics and all scales (except SIFT on Homography Estimation Accuracy and large image resolution). Notice how R2D2, DISK and SuperPoint rankings differ across resolutions.}
  \label{fig:ana-inference-input-size}
\end{figure*}

\begin{table*}
  \centering
  \footnotesize
  \begin{tabular}{c|c|cc|cc|cc|cc|cc}
 \hline
 \multicolumn{12}{c}{HPatches} \\
 \hline
 & Size & \multicolumn{2}{c|}{Repeatability} & \multicolumn{2}{c|}{Hom. Est. Acc.} & \multicolumn{2}{c|}{Hom. Est. AUC} & \multicolumn{2}{c|}{MMA} & \multicolumn{2}{c}{\# of keypoints} \\
 \hline
 & & \footnotesize{$\epsilon=1$} & \footnotesize{$\epsilon=3$} & \footnotesize{$\epsilon=1$} & \footnotesize{$\epsilon=3$} & \footnotesize{$\epsilon=1$} & \footnotesize{$\epsilon=3$} & \footnotesize{$\epsilon=1$} & \footnotesize{$\epsilon=3$} & \footnotesize{pre-match} & \footnotesize{post-match} \\
 \hline
\multirow{5}{*}{\normalsize{SuperPoint}}
& 240 & 0.39$\times$ & 0.64 & 0.47$\times$ & 0.83 & 0.20$\times$ & 0.54 & 0.48$\times$ & 0.76 & 305 & 187  \\
& 360 & 0.37 & 0.63 & 0.49 & 0.83 & 0.23 & 0.55 & 0.45 & 0.74 & 569 & 341  \\
& 480 & 0.34 & 0.61 & 0.43 & 0.80 & 0.20 & 0.51 & 0.41 & 0.72 & 847 & 499  \\
& 600 & 0.31 & 0.59 & 0.41 & 0.79 & 0.19 & 0.49 & 0.38 & 0.69 & 1141 & 655  \\
& 720 & 0.28 & 0.56\checkmark & 0.37 & 0.74\checkmark & 0.16 & 0.45\checkmark & 0.35 & 0.66\checkmark & 1460 & 813  \\
\hline\
\multirow{5}{*}{\normalsize{SIFT}}
& 240 & 0.35$\times$ & 0.54 & 0.69$\times$ & 0.89 & 0.39$\times$ & 0.68 & 0.48$\times$ & 0.57 & 677 & 298  \\
& 360 & 0.33 & 0.53 & 0.63 & 0.85 & 0.37 & 0.65 & 0.45 & 0.56 & 1331 & 570  \\
& 480 & 0.31 & 0.52 & 0.60 & 0.84 & 0.34 & 0.61 & 0.41 & 0.55 & 2189 & 910  \\
& 600 & 0.29 & 0.51 & 0.53 & 0.80 & 0.32 & 0.58 & 0.38 & 0.53 & 3212 & 1297  \\
& 720 & 0.27 & 0.50\checkmark & 0.49 & 0.75\checkmark & 0.28 & 0.53\checkmark & 0.35 & 0.50\checkmark & 4273 & 1681  \\
\hline
\multirow{5}{*}{\normalsize{R2D2}}
& 240 & 0.36$\times$ & 0.70 & 0.52$\times$ & 0.81 & 0.25$\times$ & 0.56 & 0.44$\times$ & 0.79 & 1037 & 351  \\
& 360 & 0.37 & 0.73 & 0.48 & 0.82 & 0.23 & 0.54 & 0.38 & 0.77 & 3193 & 1042  \\
& 480 & 0.36 & 0.72 & 0.45 & 0.79 & 0.20 & 0.50 & 0.34 & 0.75 & 6088 & 1967  \\
& 600 & 0.33 & 0.71 & 0.37 & 0.76 & 0.17 & 0.46 & 0.31 & 0.72 & 9517 & 2994  \\
& 720 & 0.30 & 0.68\checkmark & 0.32 & 0.69\checkmark & 0.14 & 0.42\checkmark & 0.26 & 0.67\checkmark & 12036 & 3698  \\
\hline
\multirow{5}{*}{\normalsize{DISK}}
& 240 & 0.46$\times$ & 0.72 & 0.54$\times$ & 0.86 & 0.27$\times$ & 0.60 & 0.62$\times$ & 0.87 & 841 & 484  \\
& 360 & 0.42 & 0.71 & 0.54 & 0.85 & 0.26 & 0.58 & 0.58 & 0.86 & 1847 & 1030  \\
& 480 & 0.38 & 0.69 & 0.45 & 0.80 & 0.22 & 0.52 & 0.52 & 0.84 & 3349 & 1794  \\
& 600 & 0.34 & 0.67 & 0.39 & 0.75 & 0.19 & 0.47 & 0.47 & 0.81 & 5152 & 2647  \\
& 720 & 0.31 & 0.65\checkmark & 0.34 & 0.71\checkmark & 0.16 & 0.43\checkmark & 0.42 & 0.77\checkmark & 7417 & 3732  \\
\hline
\multirow{5}{*}{\normalsize{LoFTR (outdoor)}}
& 240 & - & - & 0.74$\times$ & 0.90 & 0.44$\times$ & 0.72 & 0.76$\times$ & 0.93 & 1280 & 662  \\
& 360 & - & - & 0.71 & 0.90 & 0.42 & 0.70 & 0.70 & 0.92 & 2878 & 1533  \\
& 480 & - & - & 0.65 & 0.87 & 0.37 & 0.65 & 0.64 & 0.91 & 5109 & 2719  \\
& 600 & - & - & 0.59 & 0.83 & 0.33 & 0.61 & 0.58 & 0.89 & 7987 & 4166  \\
& 720 & - & - & 0.53 & 0.79\checkmark & 0.30 & 0.57\checkmark & 0.53 & 0.87\checkmark & 11490 & 5804  \\\cline{2-12}
\hline
\hline
\multirow{5}{*}{\normalsize{\model (top-10k)}}
& 240 & 0.76\checkmark & 0.90 & 0.80\checkmark & 0.93 & 0.54\checkmark & 0.78 & 0.73\checkmark & 0.79 & 10000 & 4816  \\
& 360 & 0.68 & 0.85 & 0.71 & 0.91 & 0.46 & 0.72 & 0.66 & 0.75 & 10000 & 4515  \\
& 480 & 0.62 & 0.81 & 0.62 & 0.87 & 0.40 & 0.66 & 0.59 & 0.71 & 10000 & 4283  \\
& 600 & 0.57 & 0.77 & 0.55 & 0.81 & 0.35 & 0.59 & 0.53 & 0.67 & 10000 & 4092  \\
& 720 & 0.53 & 0.73$\times$ & 0.49 & 0.76$\times$ & 0.30 & 0.53$\times$ & 0.48 & 0.63$\times$ & 10000 & 3945  \\
\hline
\multirow{5}{*}{\normalsize{\model (top-5k)}}
& 240 & 0.69\checkmark & 0.86 & 0.79\checkmark & 0.93 & 0.53\checkmark & 0.77 & 0.71\checkmark & 0.77 & 5000 & 2331  \\
& 360 & 0.62 & 0.80 & 0.70 & 0.90 & 0.45 & 0.70 & 0.64 & 0.73 & 5000 & 2181  \\
& 480 & 0.56 & 0.76 & 0.60 & 0.85 & 0.39 & 0.64 & 0.57 & 0.69 & 5000 & 2074  \\
& 600 & 0.51 & 0.71 & 0.52 & 0.80 & 0.33 & 0.57 & 0.52 & 0.65 & 5000 & 1983  \\
& 720 & 0.47 & 0.67$\times$ & 0.48 & 0.74$\times$ & 0.29 & 0.52$\times$ & 0.47 & 0.61$\times$ & 5000 & 1919  \\
\hline
\multirow{5}{*}{\normalsize{\model (top-1k)}}
& 240 & 0.54\checkmark & 0.73 & 0.74\checkmark & 0.91 & 0.44$\times$ & 0.72 & 0.66\checkmark & 0.71 & 1000 & 429  \\
& 360 & 0.48 & 0.66 & 0.62 & 0.86 & 0.38 & 0.65 & 0.59 & 0.67 & 1000 & 408  \\
& 480 & 0.43 & 0.61 & 0.53 & 0.81 & 0.32 & 0.58 & 0.52 & 0.63 & 1000 & 389  \\
& 600 & 0.38 & 0.57 & 0.47 & 0.75 & 0.28 & 0.52 & 0.47 & 0.59 & 1000 & 376  \\
& 720 & 0.35 & 0.53$\times$ & 0.43 & 0.69$\times$ & 0.25 & 0.48\checkmark & 0.42 & 0.55$\times$ & 1000 & 366 \\
\hline
 \end{tabular}
 \caption{\textbf{\model is the only model that benefits from running at lower resolutions.} On each metric and method, we compare the lowest (resolution=240, $\epsilon$=1) pair, to the highest (resolution=720, $\epsilon$=3) pair. Since the resolution / $\epsilon$ ratio is the same for both pairs, the measured level of accuracy is equivalent.}
 \label{tab:exp-ana-inference-size}
\end{table*}

All of the HPatches metrics reported in this paper use an $\epsilon$-distance error threshold to determine whether or not a pixel position is considered close-enough to its ground truth. A low $\epsilon$ means the metrics are reported in a highly accurate regime, while a high value of $\epsilon$ allows for some local mistakes to occur. However, $\epsilon$ is an absolute pixel distance, which means that metrics might vary in non-trivial ways as the input resolution changes during inference.

Existing methods tend to report metrics using high values of $\epsilon$. For example, D2-Net~\cite{dusmanu2019d2}, R2D2~\cite{revaud2019r2d2} and DISK~\cite{tyszkiewicz2020disk} all report MMA with $\epsilon$-thresholds up to 10. We argue this is unnecessary. Tasks that do not require accurate keypoints (i.e. high values of $\epsilon$) might want to reconsider running their keypoint model on lower resolution images (to reduce computational cost). As an input image is downscaled by a factor $\gamma$, $\epsilon$ should also be downscaled by the same factor in order to keep its relative size constant. So in theory, a metric reported on resolution $\alpha$ with error-threshold $\epsilon$ should be roughly equivalent to the same metric reported on resolution $\frac{\alpha}{\gamma}$ with error threshold $\frac{\epsilon}{\gamma}$.

In \cref{tab:exp-ana-inference-size}, we show this initial intuition is not correct. When looking at resolutions of 720 and 240 (i.e. downscaling of $\gamma$=3), existing methods all underperform their expected theoretical values. \model is the only model that consistently gain from running at lower resolutions. This is likely caused by \model's ability to obtain high number of keypoints on low resolution images while other methods are limited by their sparsity constraints (i.e. cell-based keypoint detection and NMS).

Additionally, we show in \cref{fig:ana-inference-input-size} that \model's performance against sparse keypoint methods is robust across a wide range of resolutions; with the exception of SIFT on large image resolution, using Homography Estimation Accuracy metric. One can also observe large performance gain (except on MMA) versus LoFTR~\cite{sun2021loftr}(dense keypoint method) in the low resolution regime.

%% file: suppl/experiments/topk.tex
\subsection{Increasing top-k improves \model, but saturates early}

\begin{figure}[h]
  \centering
  \includegraphics[width=\linewidth]{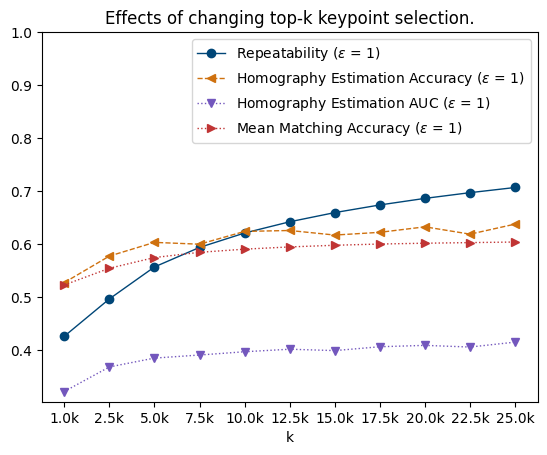}
  \caption{\textbf{Increasing top-k keypoint selection gives an initial boost in performance, but tend to get dimishing returns for $\mathbf{k > 10,000}$.}}
  \label{fig:plot-top-k}
\end{figure}

\begin{table*}
  \centering
  \footnotesize
  \begin{tabular}{c|cc|cc|cc|cc|cc}
  \hline
  \multicolumn{11}{c}{HPatches} \\
 \hline
 & \multicolumn{2}{c|}{Repeatability} & \multicolumn{2}{c|}{Hom. Est. Acc.} & \multicolumn{2}{c|}{Hom. Est. AUC} & \multicolumn{2}{c|}{MMA} & \multicolumn{2}{c}{\# of keypoints} \\
 \hline
 k & \footnotesize{$\epsilon=1$} & \footnotesize{$\epsilon=3$} & \footnotesize{$\epsilon=1$} & \footnotesize{$\epsilon=3$} & \footnotesize{$\epsilon=1$} & \footnotesize{$\epsilon=3$} & \footnotesize{$\epsilon=1$} & \footnotesize{$\epsilon=3$} & \footnotesize{pre-match} & \footnotesize{post-match} \\
 \hline
1.0k & 0.43 & 0.61 & 0.53 & 0.81 & 0.32 & 0.58 & 0.52 & 0.63 & 1000 & 389  \\
2.5k & 0.50 & 0.70 & 0.58 & 0.83 & 0.37 & 0.62 & 0.55 & 0.67 & 2500 & 1006  \\
5.0k & 0.56 & 0.76 & 0.60 & 0.85 & 0.39 & 0.64 & 0.57 & 0.69 & 5000 & 2074  \\
7.5k & 0.59 & 0.79 & 0.60 & 0.85 & 0.39 & 0.64 & 0.58 & 0.70 & 7500 & 3169  \\
10.0k & 0.62 & 0.81 & 0.62 & 0.87 & 0.40 & 0.66 & 0.59 & 0.71 & 10000 & 4283  \\
12.5k & 0.64 & 0.82 & 0.63 & 0.87 & 0.40 & 0.66 & 0.59 & 0.72 & 12500 & 5408  \\
15.0k & 0.66 & 0.83 & 0.62 & 0.86 & 0.40 & 0.66 & 0.60 & 0.72 & 15000 & 6541  \\
17.5k & 0.67 & 0.84 & 0.62 & 0.86 & 0.41 & 0.66 & 0.60 & 0.73 & 17500 & 7677  \\
20.0k & 0.69 & 0.85 & 0.63 & 0.86 & 0.41 & 0.66 & 0.60 & 0.73 & 20000 & 8821  \\
22.5k & 0.70 & 0.85 & 0.62 & 0.87 & 0.41 & 0.66 & 0.60 & 0.73 & 22500 & 9963  \\
25.0k & 0.71 & 0.86 & 0.64 & 0.87 & 0.42 & 0.67 & 0.60 & 0.73 & 25000 & 11106  \\
\hline

 \end{tabular}
 \caption{Numerical results used in \cref{fig:plot-top-k}.}
 \label{tab:exp-top-k}
\end{table*}

Increasing top-k has shown multiple times (cf. \cref{tab:exp-ana-inference-size}) to improve overall results. In this experiment, we simply vary the parameter k in order to monitor \model's performance.

As can be observed in \cref{fig:plot-top-k,tab:exp-top-k}, results start to saturate after $k=10,000$; on Homography Estimation and MMA. Repeatability continuously increases as we increase k, but that is simply a consequences of getting more keypoints (i.e. the keypoint overlaps become more likely).

%% file: suppl/experiments/dist-post-proc.tex
\subsection{Improving MMA using ratio-test or double-softmax filtering}

All previously reported results have been computed using MNN matching on unprocessed cosine distances. There are, however, known distance post-processing techniques used to reduce false positive matching. The ratio-test~\cite{lowe2004distinctive} is one of those. The distance of the best match is divided by the distance of the second best match. A low value indicates a large difference between the two best distances, which indicates a measure of relative distinctiveness. Therefore, filtering out matches with high ratio values do tend to reduce matching errors caused by repeated similar keypoints (e.g. window corners of a building).

More recently\cite{sun2021loftr}, a similar idea has emerged from the probabilistic formulation of the matching problem: Filtering out low-probability matches seems like a natural way to reduce false positive matches.

\begin{figure}[h]
\centering
\includegraphics[width=\linewidth]{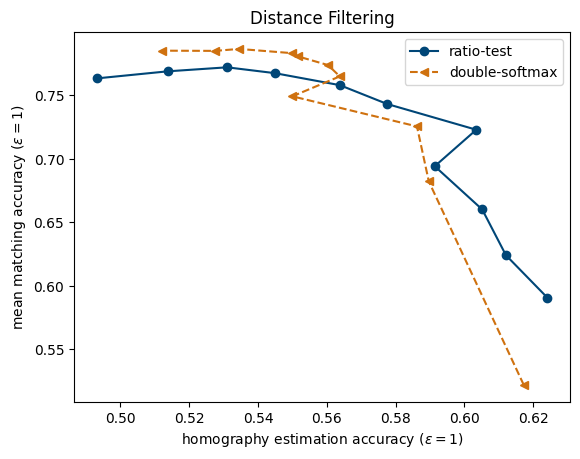}
\caption{\textbf{MMA / Homography trade-off can be controlled with ratio-test and double-softmax filtering.} Different threshold values are tested (between 0.5 and 1) using both methods.}
\label{fig:plot-dist-filtering}
\end{figure}

In \cref{fig:plot-dist-filtering,tab:exp-dist-pp}, we show that using either ratio-test or double-softmax filtering can help \model trade Homography Estimation for MMA.

\begin{table*}
  \centering
  \footnotesize
  \begin{tabular}{c|c|cc|cc|cc|cc}
  \hline
  \multicolumn{10}{c}{HPatches} \\
 \hline
 & & \multicolumn{2}{c|}{Hom. Est. Acc.} & \multicolumn{2}{c|}{Hom. Est. AUC} & \multicolumn{2}{c|}{MMA} & \multicolumn{2}{c}{\# of keypoints} \\
 \hline
 & Threshold & \footnotesize{$\epsilon=1$} & \footnotesize{$\epsilon=3$} & \footnotesize{$\epsilon=1$} & \footnotesize{$\epsilon=3$} & \footnotesize{$\epsilon=1$} & \footnotesize{$\epsilon=3$} & \footnotesize{pre-match} & \footnotesize{post-match} \\
 \hline
 \multirow{11}{*}{\normalsize{ratio-test}}
& 1.00 & 0.62 & 0.87 & 0.40 & 0.66 & 0.59 & 0.71 & 10000 & 4283 \\
& 0.95 & 0.61 & 0.87 & 0.40 & 0.65 & 0.62 & 0.75 & 10000 & 3771 \\
& 0.90 & 0.61 & 0.86 & 0.40 & 0.65 & 0.66 & 0.79 & 10000 & 3314 \\
& 0.85 & 0.59 & 0.84 & 0.40 & 0.64 & 0.69 & 0.83 & 10000 & 2926 \\
& 0.80 & 0.60 & 0.83 & 0.40 & 0.63 & 0.72 & 0.86 & 10000 & 2599 \\
& 0.75 & 0.58 & 0.82 & 0.40 & 0.62 & 0.74 & 0.88 & 10000 & 2314 \\
& 0.70 & 0.56 & 0.82 & 0.39 & 0.61 & 0.76 & 0.89 & 10000 & 2061 \\
& 0.65 & 0.54 & 0.78 & 0.39 & 0.59 & 0.77 & 0.90 & 10000 & 1830 \\
& 0.60 & 0.53 & 0.78 & 0.38 & 0.58 & 0.77 & 0.91 & 10000 & 1620 \\
& 0.55 & 0.51 & 0.75 & 0.37 & 0.57 & 0.77 & 0.91 & 10000 & 1427 \\
& 0.50 & 0.49 & 0.76 & 0.37 & 0.55 & 0.76 & 0.90 & 10000 & 1248 \\
\hline
\multirow{11}{*}{\normalsize{double-softmax}}
 & 1.00 & 0.62 & 0.86 & 0.40 & 0.65 & 0.52 & 0.63 & 10000 & 5321 \\
 & 0.95 & 0.59 & 0.86 & 0.39 & 0.64 & 0.68 & 0.83 & 10000 & 3650 \\
 & 0.90 & 0.59 & 0.84 & 0.39 & 0.63 & 0.73 & 0.88 & 10000 & 3244 \\
 & 0.85 & 0.55 & 0.83 & 0.39 & 0.62 & 0.75 & 0.91 & 10000 & 2923 \\
 & 0.80 & 0.56 & 0.81 & 0.39 & 0.61 & 0.77 & 0.92 & 10000 & 2616 \\
 & 0.75 & 0.56 & 0.80 & 0.39 & 0.61 & 0.77 & 0.92 & 10000 & 2319 \\
 & 0.70 & 0.55 & 0.79 & 0.39 & 0.60 & 0.78 & 0.93 & 10000 & 2041 \\
 & 0.65 & 0.55 & 0.79 & 0.38 & 0.59 & 0.78 & 0.93 & 10000 & 1782 \\
 & 0.60 & 0.53 & 0.78 & 0.38 & 0.58 & 0.79 & 0.93 & 10000 & 1547 \\
 & 0.55 & 0.53 & 0.76 & 0.38 & 0.57 & 0.79 & 0.92 & 10000 & 1335 \\
 & 0.50 & 0.51 & 0.76 & 0.38 & 0.57 & 0.79 & 0.92 & 10000 & 1141 \\
\hline

 \end{tabular}
 \caption{Numerical results used in \cref{fig:plot-dist-filtering}.}
 \label{tab:exp-dist-pp}
\end{table*}

%% file: suppl/experiments/padding.tex
\subsection{Use NO padding to learn keypoints.}

Zero padding is commonly used in various models. It is the process of adding a 0-filled border to an image or dense feature map. A common example is when using 3x3 convolutions, adding a padding of 1 will ensure the spatial shape of the input is preserved, otherwise it would be reduced.

The use or lack of padding is rarely mentioned by existing keypoint methods. However, we find that most implementations do in fact avoid the use of zero padding. Other implementations might use it, but then compensate by removing an arbitrarily-sized border from the dense outputs.

The reason for not using padding in \model's case is because it creates easily detectable corners and edges on the image borders, therefore causing overfitting during training.

To show the importance of \textbf{not} using padding when learning keypoints, we provide two tables (cf. \cref{tab:exp-padding}) as evidence of the adverse effects of using it.

\begin{table*}
\begin{subtable}{\textwidth}
  \centering
  \footnotesize
  \begin{tabular}{c|cc|cc|cc|cc|cc}
  \hline
  \multicolumn{11}{c}{HPatches} \\
 \hline
 & \multicolumn{2}{c|}{Repeatability} & \multicolumn{2}{c|}{Hom. Est. Acc.} & \multicolumn{2}{c|}{Hom. Est. AUC} & \multicolumn{2}{c|}{MMA} & \multicolumn{2}{c}{\# of keypoints} \\
 \hline
padding & 0.59 & 0.79 & 0.59 & 0.84 & 0.39 & 0.63 & 0.57 & 0.68 & 10000 & 4222 \\
no padding & \textbf{0.62} & \textbf{0.81} & \textbf{0.62} & \textbf{0.87} & \textbf{0.40} & \textbf{0.66} & \textbf{0.59} & \textbf{0.71} & 10000 & 4283 \\
 
 \end{tabular}
\end{subtable}
\begin{subtable}{\textwidth}
\setlength{\tabcolsep}{3.5pt}
  \centering
  \footnotesize
  \begin{tabular}{l|ccc|cc|ccc|cc|ccc|cc}
 \hline
  & \multicolumn{15}{c}{ScanNet} \\
 \hline
  & \multicolumn{5}{c|}{Rotation} & \multicolumn{5}{c|}{Translation} & \multicolumn{5}{c}{Chamfer} \\
 \hline
  & \multicolumn{3}{c|}{Accuracy $\uparrow$} & \multicolumn{2}{c|}{Error $\downarrow$} & \multicolumn{3}{c|}{Accuracy $\uparrow$} & \multicolumn{2}{c|}{Error $\downarrow$} & \multicolumn{3}{c|}{Accuracy $\uparrow$} & \multicolumn{2}{c}{Error $\downarrow$} \\
 \hline
 &  $5^{\circ}$ & $10^{\circ}$ & $45^{\circ}$ & Mean & Med. & 5 & 10 & 25 & Mean & Med. & 1 & $5$ & 10 & Mean & Med. \\
 \hline
 padding & 93.7 & 97.2 & \textbf{99.7} &  2.1 &  0.9 & 75.2 & 89.8 & 97.4 &  5.2 &  2.5 & 85.6 & 95.9 & 97.8 &  4.6 &  \textbf{0.1} \\
 no padding & \textbf{98.1} & \textbf{99.0} & 99.6 &  \textbf{1.7} & \textbf{0.8} & \textbf{82.9} & \textbf{94.8} & \textbf{99.0} & \textbf{4.1} &  \textbf{2.1} & \textbf{92.8} & \textbf{98.3} & \textbf{99.1} &  \textbf{4.3} &  \textbf{0.1} \\
  \end{tabular}
 \end{subtable}
\caption{\textbf{Avoid using padding to learn good keypoints.}}
 \label{tab:exp-padding}
\end{table*}

%% file: suppl/implementation.tex
\section{Implementation details}

\subsection{Data augmentation}

Here we detail the data augmentation used by \model during training, provided by Albumentation library (\cref{fig:augmentation}):
\begin{figure}[h]
\centering
\begin{lstlisting}[language=Python]
import albumentations as A

silk_augmentation = A.Compose([
    A.RandomGamma(
    	p=0.1, gamma_limit=(15, 65)
    ),
    A.HueSaturationValue(
    	p=0.1, val_shift_limit=(-100, -40)
    ),
    A.Blur(
    	p=0.1, blur_limit=(3, 9)
    ),
    A.MotionBlur(
    	p=0.2, blur_limit=(3, 25)
    ),
    A.RandomBrightnessContrast(
    	p=0.5, 
    	brightness_limit=(-0.3, 0.0), 
    	contrast_limit=(-0.5, 0.3)
    ),
    A.GaussNoise(p=0.5),
], p=0.95)
\end{lstlisting}
\caption{Pseudo-code: Data augmentation for \model.}
\label{fig:augmentation}
\end{figure}